\documentclass[10pt,twocolumn,letterpaper]{article}

\usepackage{iccv}              % To produce the CAMERA-READY version
\usepackage[linesnumbered,ruled,vlined]{algorithm2e}
\usepackage{tabularx}
\usepackage{multirow} 
\usepackage[accsupp]{axessibility}  % Improves PDF readability for those with disabilities.
\usepackage{multirow}
\definecolor{iccvblue}{rgb}{0.21,0.49,0.74}
\usepackage[pagebackref,breaklinks,colorlinks,allcolors=iccvblue]{hyperref}

\def\paperID{*****} % *** Enter the Paper ID here
\def\confName{ICCV}
\def\confYear{2025}

\title{A Unified Detection Pipeline for Robust Object Detection in Fisheye-Based Traffic Surveillance }

\author{
Neema Jakisa Owor$^{1}$\thanks{Equal contribution} \quad
Joshua Kofi Asamoah$^{2,3}$\footnotemark[1] \quad
Tanner Wambui Muturi$^{1}$\footnotemark[1] \quad
Anneliese Jakisa Owor$^{1}$ \\
Blessing Agyei Kyem$^{2,3}$ \quad
Andrews Danyo$^{2,3}$ \quad
Yaw Adu-Gyamfi$^{1}$ \quad
Armstrong Aboah$^{2,3}$\thanks{Corresponding author} \\
\vspace{0.3em}
$^1$University of Missouri–Columbia \quad
$^2$North Dakota State University \quad
$^3$SMART Lab \\
\vspace{0.3em}
\parbox{\textwidth}{\centering\footnotesize%
\texttt{\{nodyv, twmtyg, ajotdc, adugyamfiy\}@missouri.edu} \\
\texttt{\{joshua.asamoah, blessing.kyem, andrews.danyo, armstrong.aboah\}@ndsu.edu}%
}\\
\vspace{0.3em}
\small{$^*$First Authors} \quad
\small{$^\dagger$Corresponding author}
}

\begin{document}
\maketitle
\begin{abstract}
Fisheye cameras offer an efficient solution for wide-area traffic surveillance by capturing large fields of view from a single vantage point. However, the strong radial distortion and nonuniform resolution inherent in fisheye imagery introduce substantial challenges for standard object detectors, particularly near image boundaries where object appearance is severely degraded. In this work, we present a detection framework designed to operate robustly under these conditions. Our approach employs a simple yet effective pre and post processing pipeline that enhances detection consistency across the image, especially in regions affected by severe distortion. We train several state-of-the-art detection models on the fisheye traffic imagery and combine their outputs through an ensemble strategy to improve overall detection accuracy. Our method achieves an F1 score of \textbf{0.6366} on the 2025 AI City Challenge Track 4, placing \textbf{8th} overall out of \textbf{62 teams}. These results demonstrate the effectiveness of our framework in addressing issues inherent to fisheye imagery.
\end{abstract}    
\section{Introduction}
\label{sec:intro}

Object detection is a core component of intelligent transportation systems (ITS), supporting critical functions such as vehicle detection and classification, traffic state estimation, and incident detection \cite{zhang2022monocular}. The effectiveness of these tasks depends not only on algorithmic accuracy but also on the spatial and temporal quality of visual input \cite{ghahremannezhad2023object}. In practice, most ITS deployments rely on rectilinear cameras, whose narrow fields of view limit the ability to capture complete scenes, particularly at complex locations like intersections and merges \cite{gia2024enhancing}. To address these limitations, systems often deploy multiple cameras to extend coverage. Although this strategy improves visibility, it introduces added hardware costs, increases calibration complexity, and reduces spatial continuity across views. These trade-offs have prompted growing interest in camera systems that offer wide-area coverage from a single viewpoint, with reduced infrastructure and improved scene-level coherence.

\begin{figure}
    \centering
    \includegraphics[width=6cm]{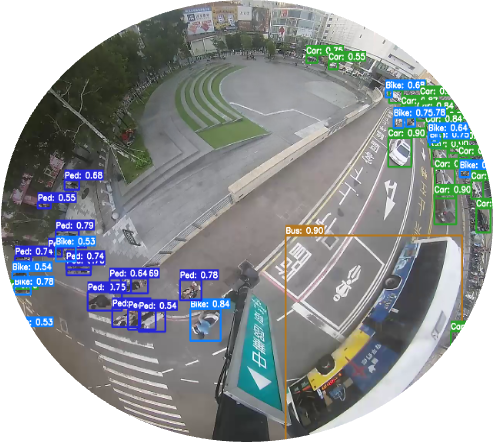}
    \caption{Fisheye-based traffic surveillance image capturing wide-area activity at an intersection using a single camera, as well as object detections from our proposed solution.}
    \label{fig:fish}
    \vspace{-1.5em}
\end{figure}

Fisheye cameras provide a promising alternative, offering panoramic fields of view up to 180 degrees from a single vantage point (Fig.~\ref{fig:fish}). This enables continuous coverage of complex scenes such as roundabouts, merges, and intersections using a single device. By capturing more of the environment in a single frame, fisheye systems enhance spatial continuity and significantly reduce hardware complexity. However, this comes at the cost of severe radial distortion, particularly near the image periphery, where objects often appear warped, compressed, or blurred \cite{gochoo2023fisheye8k, jeon2024fisheye}. In response, recent studies have reported three main approaches in addressing these limitations. First, the use of preprocessing techniques that enhance visual consistency under varying conditions. For example, methods such as illumination clustering and night-to-day conversion have been proposed to normalize image appearance prior to detection \cite{tran2024low}. Second, the introduction of distortion-aware models that aim to learn geometric invariance directly from fisheye data. These models often incorporate spherical convolutions or curved bounding boxes to account for spatial deformation without requiring explicit rectification \cite{rashed2021generalized, yu2019grid, gia2024enhancing}. Third, the use of post processing techniques that refine detection outputs through methods such as super resolution and ensemble fusion, improving localization and reducing false positives \cite{jeon2024fisheye}. While each of these approaches addresses specific limitations, they are typically applied in isolation and often involve tradeoffs in computational efficiency or latency. These limitations underscore the need for a unified detection pipeline that combines efficient preprocessing, robust model architectures, and streamlined post-processing to achieve reliable object detection on fisheye imagery.

In response to this need, we propose a unified framework designed to perform robustly under heavy radial distortion without relying on explicit geometric correction. The approach integrates a simple yet effective combination of pre and post processing steps to enhance detection consistency across the image, particularly in peripheral regions where distortion is most pronounced. Multiple state of the art object detectors are trained on fisheye traffic data, and their predictions are fused through an ensemble strategy to improve both accuracy and stability. Toward this end, the study makes the following contributions:

\begin{enumerate}
    \item We introduce a unified detection framework that operates robustly on fisheye images without explicit distortion modeling.
    \item We propose a targeted set of pre- and post-processing steps to improve detection consistency across challenging conditions. A key contribution is extending the training distribution to better handle nighttime scenes, which are underrepresented in the base fisheye dataset. To address this, we incorporate synthetic nighttime images and generate pseudo-labels using YOLO-World, which helps the model generalize more reliably under low-light conditions.
    \item We introduce a novel ensemble model that combines 4 state-of-the-art object detection models, namely, YOLOR, YOLOv12, Salience-DETR, and Co-DETR. By combining models with complementary strengths, the ensemble improves detection robustness and accuracy, particularly in the challenging visual conditions presented by fisheye traffic scenes.
    \item Finally, we conduct extensive experiments to validate the effectiveness of our proposed framework. The results demonstrate strong performance on AI City Challenge Track 4, highlighting the framework’s robustness and practical value in real-world scenarios.
\end{enumerate}

\section{Related Work}
\label{sec:review}

Object detection is central to traffic surveillance, enabling systems to track vehicles and interpret behavior over time. Advances in deep learning have significantly improved detection accuracy in conventional scenes, but fisheye imagery introduces geometric distortion that standard models struggle to handle. This section reviews core detection architectures and processing techniques, with a focus on methods developed to address challenges specific to distorted visual input.

\subsection{Object Detection for Intelligent Transportation Systems}

Deep learning-based object detectors are typically classified into two categories: two-stage and one-stage models.
\vspace{0.5em}

\noindent\textbf{Two-stage detectors} first generate region proposals and then classify objects within those regions. R-CNN \cite{girshick2014rich} introduced this pipeline, which was later improved by SPP-Net \cite{he2015spatial} and Fast R-CNN \cite{girshick2015fast} to increase efficiency and reduce redundancy. Faster R-CNN \cite{ren2016faster} further streamlined the process by integrating proposal generation into the network itself, while Mask R-CNN \cite{he2017p} extended the framework to support instance segmentation. These architectures have been widely adopted in traffic analysis tasks. For example, Zhang et al. applied Mask R-CNN to extract vehicle-level attributes such as speed, axle count, and lane position \cite{zhang2020traffic}, and Mhalla et al. developed a specialized Faster R-CNN variant for embedded multi-object traffic detection. 
% Recent work by Ali et al. \cite{ali2025small} paired Faster R-CNN with SAHI to improve detection of small objects. While two-stage models generally deliver strong performance, their multi-step nature introduces latency, making them less suitable for time-sensitive applications \cite{liu2021survey, hsu2020ratio}.

\noindent\textbf{One-stage detectors} address this issue by performing classification and localization in a single forward pass. Popular examples include YOLO \cite{redmon2016you}, SSD \cite{liu2016ssd}, and RetinaNet \cite{lin2017focal}, which have become standard for real-time applications due to their speed. Several extensions to YOLO have improved robustness under challenging conditions. For instance, Lu et al. \cite{lu2023cross} introduced a cross-scale feature fusion module and an illumination-invariant transform to improve performance under deformation and lighting variation, though challenges persist for detecting small or distant vehicles.

\noindent\textbf{Transformer-based models} have emerged as a strong alternative to CNN-based detectors. DEtection TRansformer (DETR) reformulated object detection as a set prediction task, eliminating the need for hand-crafted components like anchor boxes. Deformable DETR \cite{zhu2020deformable} addressed DETR’s slow convergence and limited resolution handling by introducing multi-scale deformable attention. Co-DETR \cite{gia2024enhancing} further enhanced encoder learning by incorporating auxiliary prediction heads. These models demonstrate state-of-the-art performance in benchmark settings but remain sensitive to input quality and often rely on additional refinement steps to maintain stability in real-world deployments\cite{BlessingAIC25,MuturiAIC25}.

\subsection{Strategies for Robust Detection under Distortion}

Improving robustness under distortion requires more than architectural changes. Preprocessing and postprocessing methods play a critical role in stabilizing detection, particularly in fisheye imagery where objects often appear warped or compressed.

\noindent\textbf{Preprocessing techniques} aim to improve input quality and reduce variability. Common strategies include region of interest selection, grayscale conversion, and inverse perspective mapping \cite{heidarizadeh2021preprocessing}. ROI selection helps reduce false positives and speeds up inference \cite{yu2018traffic, clausse2019large, zhang2020traffic}. Augmentation-based methods further improve generalization. For example, Volk et al. introduced weather-specific augmentation \cite{volk2019towards}, and Hu et al. proposed a scale-insensitive CNN with context-aware pooling \cite{hu2018sinet}. Front-end modules such as URIE \cite{son2020urie}, DeepN \cite{liu2017image}, and OWAN \cite{suganuma2019attention} preprocess distorted images before detection. Other approaches include SmoothMix \cite{lee2020smoothmix} and offline style transfer \cite{michaelis2019benchmarking}.
\vspace{0.5em}

\noindent\textbf{Postprocessing} refines raw predictions. Non-Maximum Suppression remains standard for reducing duplicates, while Weighted Boxes Fusion improves localization by averaging overlapping outputs \cite{solovyev2021weighted}. Sabater et al. introduced a similarity-based refinement method that improves accuracy on fast-moving targets with minimal cost \cite{sabater2020robust}.

% Together, these strategies help stabilize detection performance and are essential for deploying models in complex, real-time environments.

\section{Methodology}
\begin{figure}[h]
    \centering
    \includegraphics[width=8cm]{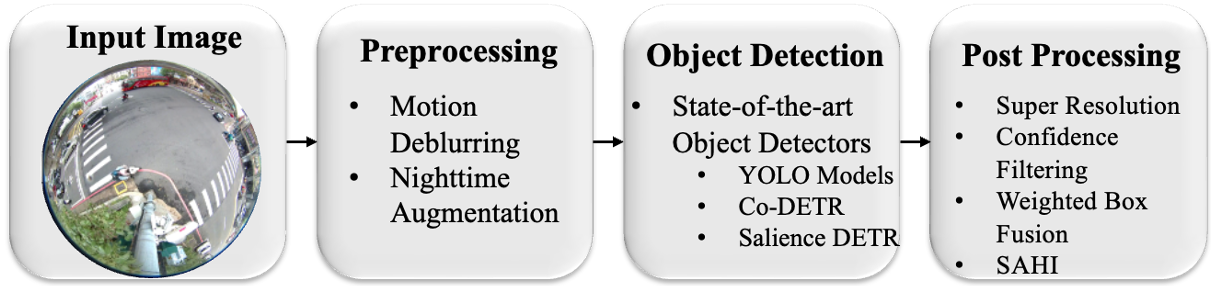}
    \caption{Overall Methodological Framework}
    \label{meth:meth}
    \vspace{-1.5em}
\end{figure}

Our proposed method is structured around three synergistic components: preprocessing, postprocessing, and efficient model ensembling. Each component is designed to target a specific set of challenges associated with object detection in fisheye surveillance imagery. Preprocessing enhances input image quality by addressing issues such as motion blur and poor illumination. Postprocessing refines model predictions through confidence filtering and resolution enhancement. Finally, a weighted ensemble strategy aggregates outputs across models for improved stability and accuracy. Our proposed framework is summarized in Figure~\ref{meth:meth}, which outlines the flow of data through each component in the system.

\subsection{Preprocessing Strategies}
\label{pre}
Two complementary preprocessing strategies were employed to enhance input quality: image deblurring and augmentation of nighttime images. These steps aim to mitigate the effects of motion blur and lighting imbalance, which significantly degrade the performance of object detectors in fisheye surveillance settings.

\subsubsection{Deblurring Motion-Affected Frames}  
\begin{figure}[h]
    \centering
    \includegraphics[width=8cm]{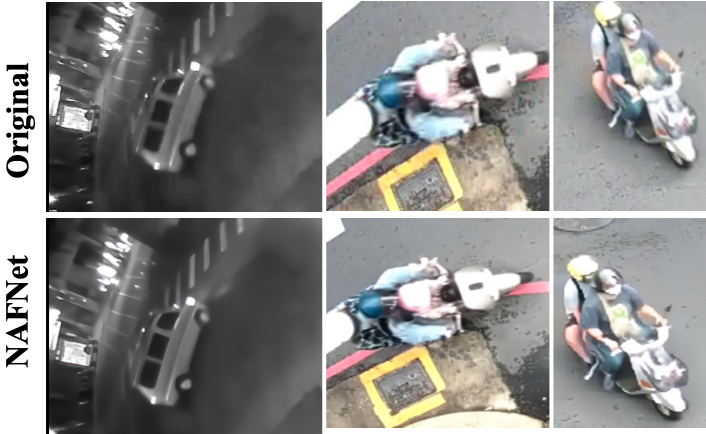}
    \caption{Showing the original images and the deblurred images}
    \label{naf:naf}
    \vspace{-1.5em}
\end{figure}

Traffic scenes captured by fisheye cameras often exhibit strong motion blur, especially near the edges of the image where vehicle motion is more tangential to the lens. To address this, we applied NAFNet\cite{chu2022nafssr}, a state-of-the-art deblurring network, to both training and testing images. This step improves feature clarity and makes it easier for detectors to localize objects that would otherwise be degraded. Figure~\ref{naf:naf} demonstrates the enhanced clarity achieved through NAFNet\cite{chu2022nafssr}, particularly in high-motion scenarios. All models were trained exclusively on deblurred images to reinforce robustness.

\subsubsection{Augmenting Nighttime Images}  
The original fisheye dataset contains relatively few nighttime scenes, limiting generalization across lighting conditions. To address this, we incorporated synthetically generated night images from a publicly available dataset\cite{pham2024improving} used in the previous year’s competition. To further enrich the dataset, YOLO-World was employed to generate pseudo-labels for the synthetic images. This augmentation strategy improved model resilience under varying illumination, enhancing performance across diverse environmental conditions.

\subsection{Post processing Strategy}
\label{post}

\subsubsection{Enhancing Small Object Visibility via Super-Resolution} 
\begin{figure}[h]
    \centering
    \includegraphics[width=8cm]{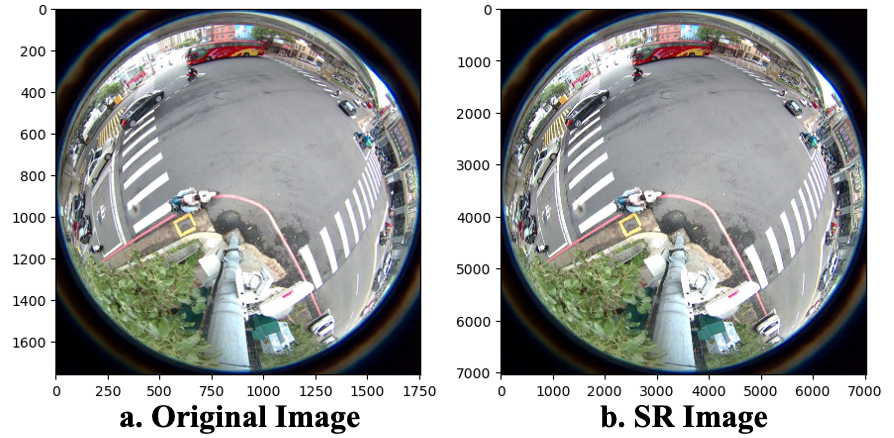}
    \caption{Showing the original image and the super-resolution image at a scale of x4}
    \label{sr:sr}
    \vspace{-1.5em}
\end{figure}

Many small objects, particularly motorcycles and pedestrians near the image boundaries, are difficult to detect due to their size and distortion. To address this, we applied super-resolution using the Dual Aggregation Transformer (DAT)\cite{chen2023dual} for Image Super-Resolution (SR). DAT increased the image resolution by a factor of four. For example, original images of size 1755×1760 were upscaled to 7020×7040. Although the upscaled images were later resized to standard input dimensions (e.g., 2520 or 1280) for inference, starting from a higher resolution preserved significantly more spatial detail. Figure~\ref{sr:sr} compares the visual richness and object clarity between original and superresoluted images, especially near the image corners.
\vspace{0.5em}

\subsubsection{Improving Small Object Precision with SAHI}To improve small object detection, we used Slicing Aided Hyper Inference (SAHI), which splits large images into overlapping slices for localized inference. Each slice was set to half the original size with 25\% overlap, and results were aggregated for final detection.

\subsubsection{Confidence-Based Filtering of Detections} 
\begin{figure}[h]
    \centering
    \includegraphics[width=8cm]{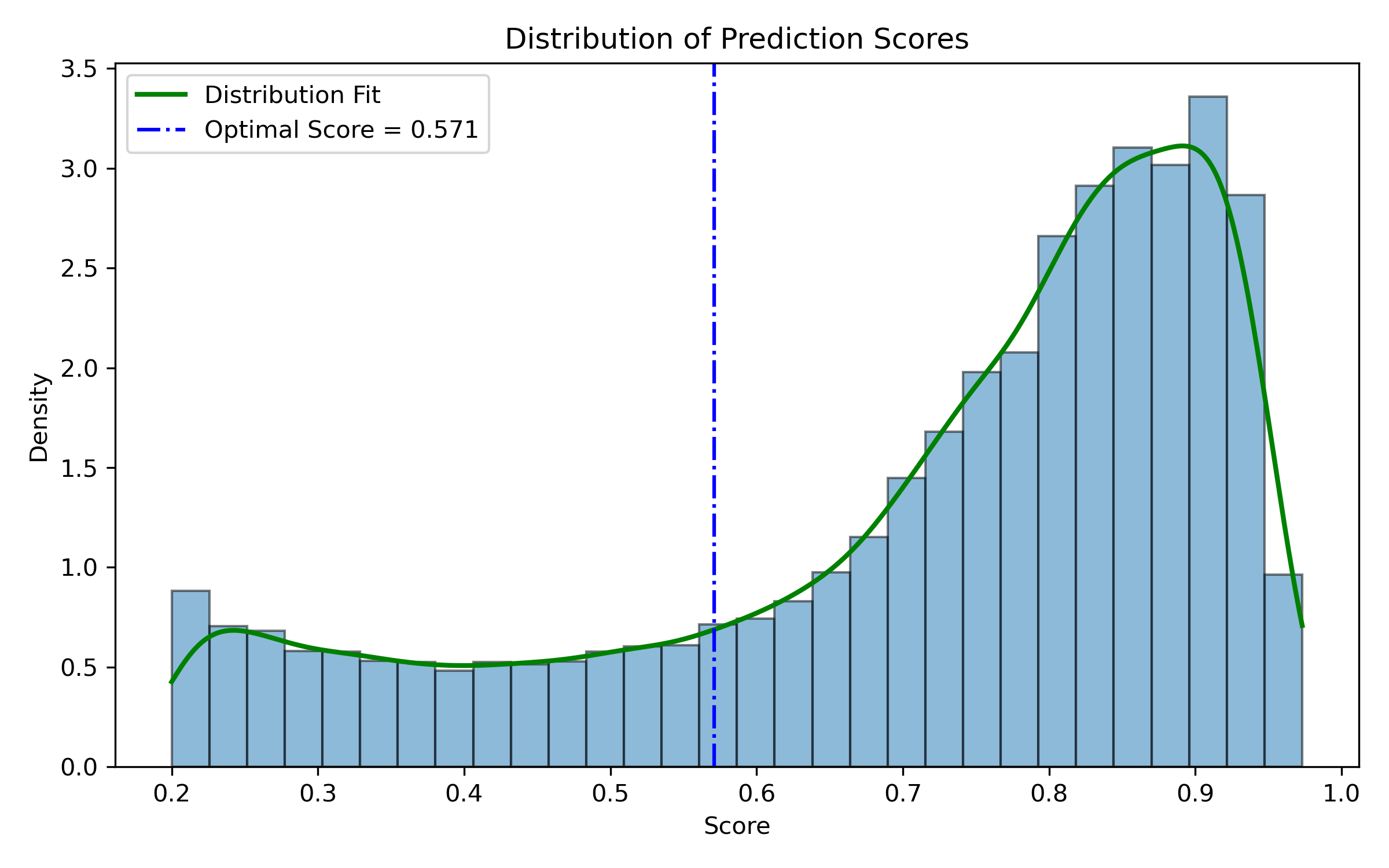}
    \caption{Distribution of predicted confidence scores}
    \label{conf:conf}
    \vspace{-1.5em}
\end{figure}

To improve detection precision, we propose a data-driven filtering strategy that uses Otsu’s\cite{goh2018performance} method to automatically determine the confidence threshold. By modeling the prediction scores as a unimodal distribution, the method selects a cutoff point that best separates low-confidence noise from high-confidence predictions. As shown in Figure~\ref{conf:conf}, the resulting threshold of 0.571 captures the high-density region of confident scores, improving overall detection precision. 

\subsubsection{Ensemble Strategy: Weighted Box Fusion}
To aggregate predictions from multiple detectors, we evaluated several fusion strategies, including Non-Maximum Suppression (NMS), Soft-NMS, Non-Maximum Weighted (NMW), and Weighted Box Fusion (WBF). While traditional NMS variants rely on heuristic suppression, they often discard useful spatial evidence from overlapping predictions. WBF addresses this limitation by computing a confidence-weighted average of box coordinates across detectors, preserving both localization precision and model agreement. This property makes WBF especially well-suited for ensemble settings where spatial alignment is consistent but confidence varies. In our experiments, WBF consistently outperformed other fusion methods and was adopted as the final step in our ensemble pipeline. The algorithmic steps are provided in Algorithm~\ref{alg:wbf}.

\begin{algorithm}[h!]
\caption{Weighted Box Fusion for Ensemble Strategy}
\label{alg:wbf}
\scriptsize
\KwIn{$\mathcal{B}$: Bounding boxes from multiple detectors,\\
\hspace{1.7em}$\mathcal{S}$: Confidence scores,\\
\hspace{1.7em}$\mathcal{L}$: Class labels,\\
\hspace{1.7em}$T$: IoU threshold for merging}
\KwOut{Fused bounding boxes $\mathcal{B}_{\text{fused}}$}
$\mathcal{B}_{\text{fused}} \leftarrow \emptyset$\;
\While{$\mathcal{B}$ not empty}{
    $b_{\text{max}} \leftarrow$ box with highest confidence\;
    $\mathcal{G} \leftarrow$ group boxes with IoU $> T$ and same label\;
    $b_{\text{fused}} \leftarrow \frac{\sum S_i \cdot b_i}{\sum S_i}$\;
    Set class and score from box with max $S_i$\;
    Add $b_{\text{fused}}$ to $\mathcal{B}_{\text{fused}}$\;
    Remove grouped boxes from $\mathcal{B}$\;
}
\Return{$\mathcal{B}_{\text{fused}}$}\;
\end{algorithm}

\vspace{-1.5em}

\section{Experiment}
\subsection{Dataset}
Track 4 of the AI City Challenge 2025 uses the Fisheye8K dataset introduced by Gochoo et al. \cite{Gochoo_2023_CVPR}, which focuses on object detection under wide-angle distortion. The dataset contains 8000 images collected from 35 fisheye cameras deployed across Hsinchu City, Taiwan. Images were captured at resolutions of 1080 by 1080 and 1280 by 1280, representing a variety of traffic scenes with strong geometric distortion. Annotations cover five object categories, including Bus, Bike, Car, Pedestrian, and Truck, with over 157,000 labeled bounding boxes. The data is split into 5400 training images and 2560 for validation. 
% Fisheye8K provides a strong testbed for evaluating detection models in environments where spatial distortion significantly impacts performance.
\vspace{0.5em}

\noindent\textbf{Challenges:} The dataset introduces a set of challenges that make object detection particularly difficult without additional processing. Many frames suffer from motion blur due to fast-moving vehicles captured with low shutter speeds, which makes it hard for models to learn sharp, discriminative features. There is also a clear imbalance in lighting conditions, nighttime scenes are underrepresented, which limits the model’s ability to generalize to low-light environments. Another issue is object scale and distortion. Small objects like pedestrians and motorcycles often appear near the image edges, where fisheye distortion is strongest, making them easy to miss. On top of that, the raw detector outputs can be noisy, with low-confidence predictions cluttering the final results. To address these challenges, we implemented the preprocessing and post-processing strategies discussed in Sections ~\ref{pre} and ~\ref{post}.

\subsection{Evaluation Metrics}

The submission portal reports two main evaluation metrics: mean Average Precision (mAP) and F1 score.
\vspace{0.3em}

\noindent\textbf{Mean Average Precision (mAP)} is computed as the mean of average precision over all classes:

\begin{equation}
\text{mAP} = \frac{1}{n} \sum_{k=1}^{n} \text{AP}_k
\end{equation}

where \( \text{AP}_k \) is the average precision for class \( k \), and \( n \) is the number of classes. While widely used, mAP may favor predictions with high confidence but many false positives.
\vspace{0.3em}

\noindent\textbf{F1 Score} is used as the primary ranking metric. It is the harmonic mean of Precision and Recall:

\begin{equation}
\text{F1} = \frac{2 \cdot \text{Precision} \cdot \text{Recall}}{\text{Precision} + \text{Recall}}
\end{equation}

This metric provides a more balanced view of detection quality by penalizing both false positives and false negatives.

\subsection{Implementation Details}
All models were trained on NVIDIA L40 GPUs with 48 GB of memory. To ensure a comprehensive evaluation, we experimented with both lightweight and heavyweight architectures across diverse configurations. The following outlines the implementation settings for each model included in our study.
\vspace{0.5em}

\noindent\textbf{YOLOR: }We fine-tune the COCO-pretrained YOLOR-D6 model \cite{wang2021you} on the Fisheye8K training set for 300 epochs. The training uses the Adam optimizer with a learning rate of 0.01, decay rate of 0.001, and an input size of 1280×1280. YOLOR is designed for multitask learning, leveraging both implicit and explicit knowledge representations.
\vspace{0.5em}

\noindent\textbf{YOLOv11: }For additional comparison, lightweight variants of the YOLOv11 model \cite{khanam2024yolov11} were trained on the Fisheye8K dataset using the stochastic gradient descent (SGD) optimizer with a fixed learning rate of 0.01. Specifically, we evaluate YOLOv11n and YOLOv11s at two different input resolutions: 640×640 and 1280×1280. The architectural design includes modules such as C3k2, SPPF, and C2PSA, which enhance the model’s capacity for hierarchical feature extraction and contextual representation, contributing to improved detection accuracy across varied visual scenes.
\vspace{0.5em}

\noindent\textbf{YOLOv12: } Building on these lightweight comparisons, we train four configurations of YOLOv12 \cite{tian2025yolov12}: YOLOv12n at 640 and 1280 resolutions, and YOLOv12s at 640 and 1280. Each model is trained for 300 epochs using SGD with a cosine annealing learning rate schedule starting at 0.01. The architecture includes Area Attention (A²), R-ELAN blocks, and a simplified detection head, designed to balance speed and accuracy. All models are trained on the Fisheye8K dataset.
\vspace{0.5em}

\noindent\textbf{Salience DETR: }As part of the heavyweight models in our evaluation, we include Salience DETR \cite{hou2024salience}, a transformer based architecture designed for high capacity visual representation learning. To evaluate architectural robustness across different backbone types, we fine-tune five variants of Salience DETR \cite{hou2024salience}: Swin-L, ConvNeXt, ResNet-50, FocalNet, and ResNet-5scale. All models use COCO-pretrained weights and are trained on the Fisheye8K dataset for 300 epochs with the AdamW optimizer (learning rate = 1e-4). This design introduces hierarchical salience filtering to address query redundancy and enhance multi-scale representation learning.
\vspace{0.5em}

\noindent\textbf{CO-DETR: }To further explore high-capacity transformer-based architectures, we evaluate two variants of CO DETR \cite{zong2023detrs}: one with a VIT-L backbone and another with a Co-Deformable ResNet-50 (co-defor\_res50). Both models are pretrained on COCO and fine-tuned on the Fisheye8K dataset. Training is run for 300 epochs using the AdamW optimizer with a learning rate of 1e-5.  CO-DETR enhances encoder learning with parallel auxiliary heads supervised via one-to-many label assignments, enabling improved convergence and performance.

% \noindent\textbf{YOLOR}: network structure is specifically designed for multitasking with its concept based on encoding explicit and implicit knowledge \cite{wang2021you}.

% \noindent\textbf{YOLOv11}: incorporates novel elements such as the C3k2 block, SPPF, and C2PSA that contribute to more effective feature extraction and processing. These enhancements enable the model to more effectively analyse and interpret complex visual information, potentially leading to enhanced detection accuracy across various scenarios \cite{khanam2024yolov11}.

% \noindent\textbf{YOLOv12}: At the time of writing this paper, YOLOv12 is said to surpass all popular real-time object detection in accuracy and with competitive speed \cite{tian2025yolov12}. The architecture of YOLOv12 integrates Area Attention (A²) modules, Residual Efficient Layer Aggregation Networks (R-ELAN) blocks, and a streamlined detection head, optimizing visual information processing while maintaining high accuracy.

% \noindent\textbf{Salience DETR}: enhances Detection Transformer with Hierarchical Salience filtering refinement that addresses both scale bias and query redundancy \cite{hou2024salience}. We use a COCO-pretrained Salience DETR on the dataset. 

% \noindent\textbf{Co-DETR}: enhances the encoder's ability in end-to-end detectors by training multiple parallel auxiliary heads supervised by one-to-many label assignments. We employ the CO-DETR model with the VIT-L backbone architecture, which was pretrained with the COCO \cite{zong2023detrs}.
\section{Results and Discussion}

\subsection{Quantitative Results}
We evaluate both lightweight and heavyweight models to understand their trade-offs in detection accuracy, as shown in Table~\ref{tab:recall_precision_table}. Among the lightweight models, YOLOv12s (1280) achieves the best overall performance, particularly for cars (0.597 recall, 0.810 precision) and bikes (0.472 recall, 0.766 precision), outperforming other light variants. For heavyweight models, YOLOR achieves the highest performance across nearly all categories, with standout recall and precision on bikes (0.898, 0.973), cars (0.937, 0.987), and trucks (0.970, 0.985). Given its superior accuracy and balanced precision-recall across diverse categories, we adopt YOLOR as the base model in our final ensemble framework.

\begin{table*}[h!]
\centering
\scriptsize
\caption{Recall and Precision for each model across object classes on the Validation Dataset}
\renewcommand{\arraystretch}{1.1}
\begin{tabular}{c|lc|c|cc|cc|cc|cc|cc}
\toprule
\textbf{Type} & \textbf{Model} & \textbf{Size} 
& \multicolumn{2}{c}{\textbf{Bus}} 
& \multicolumn{2}{c}{\textbf{Bike}} 
& \multicolumn{2}{c}{\textbf{Car}} 
& \multicolumn{2}{c}{\textbf{Ped}} 
& \multicolumn{2}{c}{\textbf{Truck}} \\
\cmidrule(lr){4-5} \cmidrule(lr){6-7} \cmidrule(lr){8-9} 
\cmidrule(lr){10-11} \cmidrule(lr){12-13}
& & 
& \textbf{Rec.} & \textbf{Prec.} 
& \textbf{Rec.} & \textbf{Prec.} 
& \textbf{Rec.} & \textbf{Prec.} 
& \textbf{Rec.} & \textbf{Prec.} 
& \textbf{Rec.} & \textbf{Prec.} \\
\midrule\
\multirow{8}{*}{\textbf{Lightweight}} 
& YOLOv11n & 640     & 0.426 & 0.239 & 0.455 & 0.457 & 0.397 & 0.638 & 0.071 & 0.245 & 0.156 & 0.441 \\
& YOLOv11n & 1280    & 0.475 & 0.481 & 0.376 & 0.722 & 0.563 & 0.787 & 0.201 & \textbf{0.551} & 0.313 & 0.563 \\
& YOLOv11s & 640    & 0.559 & 0.581 & 0.455 & 0.693 & 0.555 & 0.867 & 0.149 & 0.473 & 0.306 & 0.703 \\
& YOLOv11s & 1280    & 0.570 & 0.712 & 0.384 & 0.714 & \textbf{0.613} & 0.845 & 0.245 & 0.320 & \textbf{0.349} & \textbf{0.745} \\
& YOLOv12n & 640     & 0.486 & 0.503 & 0.398 & 0.772 & 0.488 & 0.786 & 0.116 & 0.281 & 0.303 & 0.347 \\
& YOLOv12n & 1280    & \textbf{0.632} & \textbf{0.708} & 0.418 & 0.648 & 0.567 & 0.733 & 0.161 & 0.363 & 0.307 & 0.679 \\
& YOLOv12s & 640    & 0.523 & 0.377 & 0.410 & 0.705 & 0.522 & \textbf{0.824} & 0.139 & 0.487 & 0.307 & 0.605 \\
& YOLOv12s & 1280    & 0.542 & 0.611 & \textbf{0.472} & \textbf{0.766} & 0.597 & 0.810 & \textbf{0.283} & 0.440 & 0.315 & 0.616 \\
\midrule
\multirow{8}{*}{\textbf{Heavyweight}} 
& Salience-DETR (Swin-L)        & 1280 & 0.928 & 0.711 & 0.774 & 0.568 & 0.833 & 0.645 & 0.728 & 0.360 & 0.612 & 0.359 \\
& Salience-DETR (ConvNeXt)      & 1280 & 0.929 & 0.671 & 0.790 & 0.608 & 0.881 & 0.709 & \textbf{0.773} & 0.354 & 0.585 & 0.416 \\
& Salience-DETR (FocalNet)      & 1280 & 0.928 & 0.706 & 0.842 & 0.636 & 0.886 & 0.723 & 0.708 & 0.367 & 0.801 & 0.436 \\
& Salience-DETR (ResNet-5scale) & 1280 & 0.808 & 0.522 & 0.765 & 0.571 & 0.791 & 0.665 & 0.682 & 0.268 & 0.489 & 0.331 \\
& Salience-DETR (ResNet-50)    & 1280 & 0.865 & 0.433 & 0.700 & 0.509 & 0.849 & 0.606 & 0.606 & 0.202 & 0.794 & 0.421 \\
& CO-DETR (VIT-L)    & 1280 & 0.927    & 0.861   & 0.693   & 0.859    & 0.869    & 0.866    & 0.546    & 0.696    & 0.837    & 0.830 \\
& CO-DETR (Co-
Deformable ResNet-50)    & 1280 & 0.905    & 0.852   & 0.674   & 0.849    & 0.860    & 0.851    & 0.522    & 0.704    & 0.817    & 0.738 \\
    & YOLOR    & 1280 & \textbf{0.933}    & \textbf{0.871}    & \textbf{0.898}    & \textbf{0.973}    & \textbf{0.937}    & \textbf{0.987}    & 0.719    & \textbf{0.974}    & \textbf{0.970}   & \textbf{0.985} \\
\bottomrule
\end{tabular}
\label{tab:recall_precision_table}
\end{table*}

% \begin{table*}[h]
% \centering
% \caption{Ablation Study of Model Components}
% \renewcommand{\arraystretch}{1.1}
% \scriptsize
% \begin{tabular}{c|l|c}
% \toprule
% \textbf{Model} & \textbf{Configuration} & \textbf{F1 Score} \\
% \midrule
% 1 & YOLOR ensemble (1280, 1536, 1920) & 0.6081 \\
% 2 & YOLOR, YOLOv12, Salience-DETR, Co-DETR & 0.6144 \\
% 3 & Model 2 + YOLOR (SR) & 0.6225 \\
% \midrule
% 4 (Ours) & Model 3 + Co-DETR (SAHI) & \textbf{0.6366} \\
% \bottomrule
% \end{tabular}
% \label{tab:ablation}
% \end{table*}

\begin{table}[h]
\centering
\caption{Ablation Study of Model Components}
\renewcommand{\arraystretch}{1.1}
\scriptsize
\begin{tabular}{c|p{5.5cm}|c}
\toprule
\textbf{Model} & \textbf{Configuration} & \textbf{F1 Score} \\
\midrule
1 & YOLOR ensemble (1280, 1536, 1920) & 0.6081 \\
2 & Model 1 + YOLOv12, Salience-DETR, Co-DETR & 0.6144 \\
3 & Model 2 + YOLOR (SR) & 0.6225 \\
\midrule
4 \textbf{(Ours)} & \textbf{Model 3 + Co-DETR (SAHI)} & \textbf{0.6366} \\
\bottomrule
\end{tabular}
\label{tab:ablation}
\end{table}

\begin{table}[h]
\centering
\caption{Top 10 Teams and Their F1 Scores}
\begin{tabular}{|c|l|c|}
\hline
\textbf{Rank} & \textbf{Team Name} & \textbf{F1 Score} \\
\hline
1 & UIT-OpenCubee & 0.6493 \\
2 & UT\_T1 & 0.6413 \\
3 & UT\_KAO & 0.6413 \\
4 & Zacian & 0.6405 \\
5 & SKKU-AutoLab & 0.6397 \\
6 & UT\_Liu & 0.6393 \\
7 & ttsc & 0.6381 \\
8 & \textbf{Smart Lab (Ours)} & \textbf{0.6366} \\
9 & MIZSU & 0.6360 \\
10 & SmartVision & 0.6342 \\
\hline
\end{tabular}
\label{tab:f1_scores}
\end{table}

\subsection{Ablation Study}

We analyze the contribution of each ensemble component through a structured ablation study, as shown in Table~\ref{tab:ablation}. The first model combines three YOLOR detectors at different input resolutions, establishing a strong baseline with an F1 score of 0.6081. To expand architectural diversity, Model 2 integrates YOLOv12, Salience-DETR, and Co-DETR, resulting in a modest but meaningful gain to 0.6144. This highlights the benefit of combining detectors with different design priors. Model 3 extends this setup by incorporating super-resolved inputs for YOLOR, yielding improved performance on fine-scale objects and increasing the score to 0.6225. Finally, Model 4 introduces spatial-aware inference using SAHI with Co-DETR, pushing the performance to 0.6366, the highest across all configurations. These results suggest that no single component is solely responsible for the gain; rather, performance emerges from the interplay between complementary modules.

% \begin{table*}[h]
% \centering
% \caption{Ablation Study of Model Components}
% \renewcommand{\arraystretch}{1.1}
% \scriptsize
% \begin{tabular}{c|l|c}
% \toprule
% \textbf{Model} & \textbf{Configuration} & \textbf{F1 Score} \\
% \midrule
% 1 & YOLOR ensemble (1280, 1536, 1920) & 0.6081 \\
% 2 & YOLOR, YOLOv12, Salience-DETR, Co-DETR & 0.6144 \\
% 3 & Model 2 + YOLOR (SR) & 0.6225 \\
% \midrule
% 4 (Ours) & Model 3 + Co-DETR (SAHI) & \textbf{0.6366} \\
% \bottomrule
% \end{tabular}
% \label{tab:ablation}
% \end{table*}

\subsection{Experimental Test Dataset}

Table~\ref{tab:f1_scores} summarizes the final results from the 2025 AI City Challenge Track 4, evaluated using the F1 Score over the full test set. Our method achieved an \textbf{F1 Score of 0.6366}, placing \textbf{8th among 62 teams}. The leading team reached a score of 0.6493, with a marginal difference of just \textbf{0.0127}, underscoring the competitiveness of our method. These findings reinforce the effectiveness of our unified pipeline in handling wide-angle distortion, improving recall of small and peripheral objects, and delivering consistent detection across complex urban scenes.

% Figure~\ref{fig:qualitative} provides qualitative results comparing model predictions against ground truth annotations. Our detector consistently recovers labeled objects and identifies additional instances, particularly along the distorted periphery of fisheye images. 

% \begin{table}[h]
% \centering
% \caption{Top 10 Teams and Their F1 Scores}
% \begin{tabular}{|c|l|c|}
% \hline
% \textbf{Rank} & \textbf{Team Name} & \textbf{F1 Score} \\
% \hline
% 1 & UIT-OpenCubee & 0.6493 \\
% 2 & UT\_T1 & 0.6413 \\
% 3 & UT\_KAO & 0.6413 \\
% 4 & Zacian & 0.6405 \\
% 5 & SKKU-AutoLab & 0.6397 \\
% 6 & UT\_Liu & 0.6393 \\
% 7 & ttsc & 0.6381 \\
% 8 & \textbf{Smart Lab (Ours)} & \textbf{0.6366} \\
% 9 & MIZSU & 0.6360 \\
% 10 & SmartVision & 0.6342 \\
% \hline
% \end{tabular}
% \label{tab:f1_scores}
% \end{table}

\section{Conclusion}

In this work, we present a unified object detection framework designed to address the unique challenges posed by fisheye imagery, including radial distortion, peripheral compression, and low-light variability. Our approach combines three core components that operate in synergy: a data-centric preprocessing pipeline, a strategically curated model ensemble, and an efficient postprocessing module.

When evaluated on the 2025 AI City Challenge Track 4, our method achieved an \textbf{F1-score of 0.6366}, placing \textbf{8th on the final leaderboard}. These results highlight the value of integrating architectural diversity and inference refinement to ensure reliable detection under wide-angle distortion. Overall, the proposed framework provides a scalable basis for advancing panoramic perception and long-range visual understanding in complex environments.

{
    \small
    \bibliographystyle{ieeenat_fullname}
    \bibliography{main}
}

% WARNING: do not forget to delete the supplementary pages from your submission 
% \input{sec/X_suppl}

\end{document}